\documentclass{article}

\PassOptionsToPackage{numbers,sort&compress}{natbib}

\usepackage[preprint]{neurips_2025}

\usepackage[dvipsnames]{xcolor}
\usepackage{hyperref}
\usepackage[utf8]{inputenc} 
\usepackage[T1]{fontenc}    
\usepackage{hyperref}       
\usepackage{url}            
\usepackage{booktabs}       
\usepackage{amsfonts}       
\usepackage{nicefrac}       
\usepackage{microtype}      
\usepackage{xcolor}         
\usepackage{amsmath, amssymb}
\DeclareMathOperator*{\argmin}{arg\,min}
\usepackage{setspace}
\usepackage{algpseudocode}  
\usepackage{float} %
\usepackage{array} %
\usepackage{multirow}
\usepackage{wrapfig}
\usepackage{varwidth} 
\usepackage[ruled,vlined]{algorithm2e}
\title{PMQ-VE: Progressive Multi-Frame Quantization for Video Enhancement}
\usepackage{enumitem}
\setlist{leftmargin=3mm} 
\usepackage[dvipsnames]{xcolor}
\definecolor{citecolor}{RGB}{119,185,0}
\definecolor{citecolor1}{RGB}{66,168,235}
\usepackage{hyperref}
\hypersetup{colorlinks=true,citecolor=citecolor1,linkcolor=BrickRed,urlcolor=Thistle}
\usepackage{graphicx}
\usepackage{makecell} 
\newcommand{\best}[1]{{{\textbf{#1}}}}
\newcommand{\second}[1]{{\underline{{#1}}}}
\usepackage{pifont} 
\usepackage{caption}
\usepackage{bm}
\author{
  \textbf{ZhanFeng Feng}$^{1\dagger}$,\enspace
  \textbf{Long Peng}$^{1\dagger\ddagger}$,\enspace
  \textbf{Xin Di}$^{1\dagger}$,\enspace
  \textbf{Yong Guo}$^{2}$,\enspace
  \textbf{Wenbo Li}$^{3}$,\enspace
  \textbf{Yulun Zhang}$^{4}$,\\
  \textbf{Renjing Pei}$^{5*}$,\enspace
  \textbf{Yang Wang}$^{1,6}$\thanks{%
    * Renjing Pei and Yang Wang are the corresponding authors. 
    † ZhanFeng Feng, Long Peng, and Xin Di contributed equally to this work. 
    ‡ Long Peng is the project leader.
  },\enspace
  \textbf{Yang Cao}$^{1}$,\enspace
  \textbf{Zheng-Jun Zha}$^{1}$\\[1ex]
  \textsuperscript{1}USTC,\enspace
  \textsuperscript{2}Max Planck Institute,\enspace
  \textsuperscript{3}CUHK,\enspace
  \textsuperscript{4}SJTU,\\
  \textsuperscript{5}Institute of Automation, Chinese Academy of Sciences,\enspace
  \textsuperscript{6}Chang’an University\\[1ex]
  {\small
    \{xiaobigfeng, longp2001, dx9826\}@mail.ustc.edu.cn,\enspace
    ywang120@chd.edu.cn
  }
  \vspace{-6mm}
}

\begin{document}

\maketitle

\begin{abstract}
Multi-frame video enhancement tasks aim to improve the spatial and temporal resolution and quality of video sequences by leveraging temporal information from multiple frames, which are widely used in streaming video processing, surveillance, and generation. Although numerous Transformer-based enhancement methods have achieved impressive performance, their computational and memory demands hinder deployment on edge devices. Quantization offers a practical solution by reducing the bit-width of weights and activations to improve efficiency. However, directly applying existing quantization methods to video enhancement tasks often leads to significant performance degradation and loss of fine details. This stems from two limitations: (a) inability to allocate varying representational capacity across frames, which results in suboptimal dynamic range adaptation; (b) over-reliance on full-precision teachers, which limits the learning of low-bit student models. To tackle these challenges, we propose a novel quantization method for video enhancement: Progressive Multi-Frame Quantization for Video Enhancement (PMQ-VE). This framework features a coarse-to-fine two-stage process: Backtracking-based Multi-Frame Quantization (BMFQ) and Progressive Multi-Teacher Distillation (PMTD). BMFQ utilizes a percentile-based initialization and iterative search with pruning and backtracking for robust clipping bounds. PMTD employs a progressive distillation strategy with both full-precision and multiple high-bit (INT) teachers to enhance low-bit models' capacity and quality. Extensive experiments demonstrate that our method outperforms existing approaches, achieving state-of-the-art performance across multiple tasks and benchmarks.The code will be made publicly available at: \url{https://github.com/xiaoBIGfeng/PMQ-VE}. 
\end{abstract}

\section{Introduction}
\begin{figure}[t]
\centering
\includegraphics[width=1.0\textwidth]{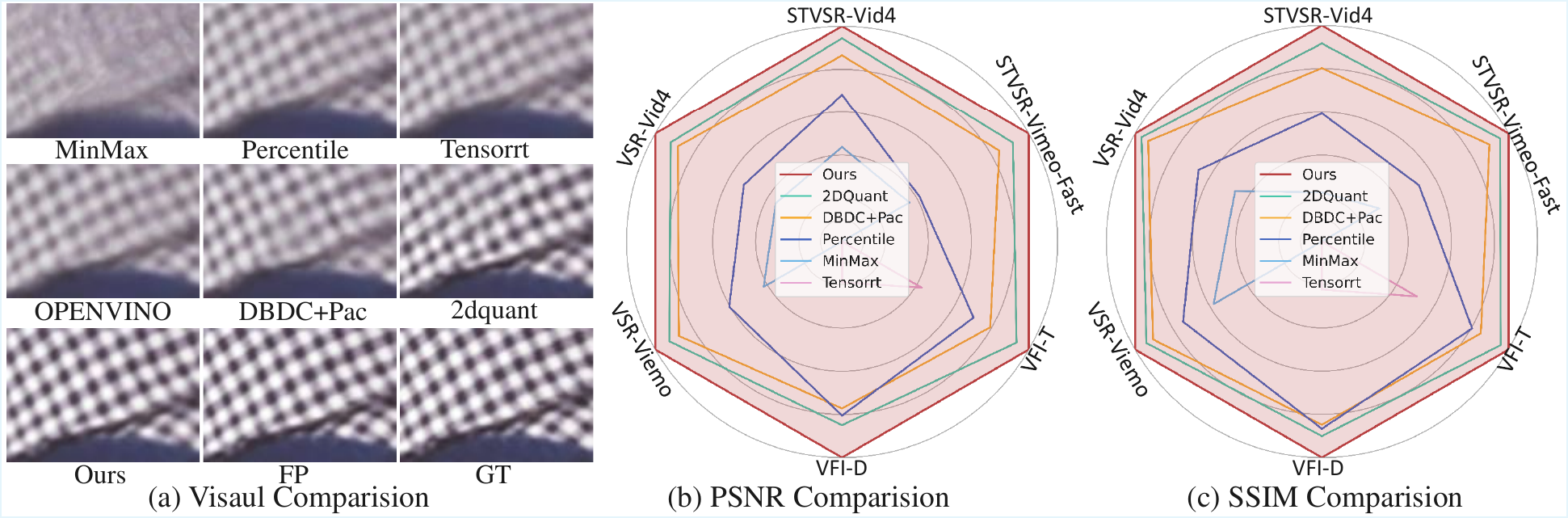}
\vspace{-4mm}
\caption{(a) Qualitative comparison of reconstructed frames using different quantization methods. Quantitative comparison of PSNR(b) and SSIM(c) improvements across three video enhancement tasks (STVSR, VFI, VSR). Our method consistently outperforms existing quantization approaches in both visual quality and quantitative metrics.}
\label{fig:fig0}
\vspace{-4mm}
\end{figure}

Multi-frame video enhancement tasks aim to enhance the spatial and temporal resolution and quality of video sequences by exploiting temporal information from multiple frames. Among these, Video Frame Interpolation (VFI) ~\cite{VFI:mamba,VFI:zhu2024dual,VFI_trans_liu2024sparse,VFI_trans_lu2022video,VFI_trans_park2023biformer,VFI_trans_zhang2023extracting,VFI:clearer,VFI:depth,VFI:ema-vfi,VFI:huang2022real,VFI:liu2017video,VFI:Neighbor,VFI:niklaus2018context}, Video Super-Resolution (VSR)~\citep{caballero2017real,cao2021video,chan2021basicvsr,chan2022basicvsr++,choi2020channel,isobe2020video,jo2018deep,peng2025boosting}, and Spatio-Temporal Video Super-Resolution (STVSR)~\cite{STSVR:zoomingslowmo,STVSR:benefit,STVSR:RSTT,STVSR:STAR,STVSR:stdan,STVSR:TMNet,STVSR:Vid4,STVSR:Vimeo90k} are the most representative video enhancement methods. They are widely employed as post-processing and pre-processing techniques in social media platforms, gaming environments, and various video perception and generation tasks ~\citep{cao2021video, STVSR:RSTT, di2024qmambabsr, peng2025directing, peng2025pixel}. Recent Transformer-based approaches for video enhancement exploit attention mechanisms to capture temporal dependencies across multiple frames, enabling substantial improvements in visual quality and structural fidelity. However, their high computational and memory demands remain a major obstacle for real-world deployment.
Therefore, numerous studies have proposed various model quantization methods to compress the bit-width of weights and activations from 32 bits (FP32) to 8, 4, or 2 bits~\cite{liu2025condiquant,liu2023noisyquant,2dquant,hong2022daq,yuan2022ptq4vit,peng2024unveiling,peng2024towards,li2020pams}. This is a crucial step in practical deployment, reducing memory consumption and inference latency. For example, PAMS~\citep{li2020pams}, a classic method in Post-Training Quantization (PTQ), introduces trainable scale parameters to dynamically learn the maximum value of the quantization range. Liu \textit{et al.} propose 2DQuant~\cite{2dquant}, a dual-stage method for image super-resolution which designs a differentiated search strategy and uses knowledge distillation~\citep{gou2021knowledge} to guide the learning of the quantization range.
However, to the best of our knowledge, exploring model quantization in video enhancement remains largely uncharted. Directly applying existing quantization methods can lead to significant issues such as performance degradation and loss of fine details. Through observations and statistical experiments, we attribute these problems to two key limitations: (a) Video enhancement models need to aggregate texture and motion information from multiple frames, leading to inter-frame differential perception of information, manifesting as differentiated activation value distributions across frames as shown in Figure~\ref{fig:fig1}(a). However, traditional quantization methods fails to allocate inconsistent representational capacity to different frames, resulting in discrepancies in dynamic range across frames, as shown in Figure~\ref{fig:fig1}(b). This results in suboptimal utilization of sub-pixel spatial details, thereby limiting reconstruction performance. (b) Quantization inevitably reduces the representational capacity of the video model. Traditional methods overlook the capacity differences between the teacher (FP32) and student models (2bit, 4bit), relying solely on full-precision teachers for distillation. This makes it challenging for the student to learn high-quality mappings given its limited capacity, resulting in difficulty when directly quantizing a high-precision network into a low-precision one, as shown in Figure~\ref{fig:fig0}(a).
To address these limitations, we propose a novel quantization framework: Progressive Multi-Frame Quantization for multi-frame Video Enhancement, called PMQ-VE. Specifically, PMQ-VE introduces a coarse-to-fine two-stage process, which includes Backtracking-based Multi-Frame Quantization (BMFQ) and Progressive Multi-Teacher Distillation (PMTD). \\
\textbf{Coarse Stage: Backtracking-based Multi-Frame Quantization.} Existing methods~\citep{2dquant,DbdcPac} typically initialize quantization bounds using the global minimum and maximum values and symmetrically shrink them inward, ignoring inter-frame variations and the asymmetric nature of distributions. To address this, as illustrated in Figure~\ref{fig:fig1}(c), BMFQ assigns frame-specific clipping bounds to better match the heterogeneous activation statistics across video frames. BMFQ employs a percentile-based initialization to suppress outliers and performs a backtracking-based search with pruning and backtracking to search the bounds efficiently. This strategy enables accurate, adaptive quantization with negligible overhead, \\
\textbf{Fine Stage: Progressive Multi-Teacher Distillation.} In the fine stage, we introduce a Progressive Multi-Teacher Distillation framework to restore the model's representational capacity under low-bit quantization. Specifically, a full-precision teacher provides fine-grained feature supervision, while an intermediate 8/4-bit teacher offers quantization-aware guidance, helping the 4/2-bit student model learn stable and informative representations under low-bit constraints, bridging the gap between the quantized and full-precision models.\\
Extensive experiments on three representative video enhancement tasks—STVSR, VSR, and VFI—demonstrate that our method achieves state-of-the-art performance across multiple benchmarks for various tasks. Our method consistently outperforms existing methods, achieving the best performance on PSNR, SSIM across various bit-width settings, as shown in Figure~\ref{fig:fig0}(b-c). The contributions of this paper can be summarized as follows: 
\vspace{-2mm}
\enlargethispage{-15mm}
\begin{itemize}[leftmargin=3mm,itemsep=0pt, topsep=0pt]
\item To the best of our knowledge, we are the first to explore model quantization in multi-frame video enhancement tasks. We propose PMQ-VE, a novel per-frame coarse-to-fine quantization framework for multi-frame video enhancement models.
\item In the coarse stage, we propose BMFQ to search quantization bounds via iterative backtracking with pruning, achieving efficient initialization. In the fine stage, we propose PMTD to leverage the knowledge of multi-level teachers to help a low-bit student model, enhancing its mapping quality and performance.
\item Extensive experiments on three video enhancement tasks (STVSR, VSR, and VFI) demonstrate that our method achieves state-of-the-art results across various benchmarks under different low-bit quantization settings, highlighting the superiority and practicability of our approach.
\end{itemize}

\section{Related work}
\subsection{Video Enhancement}
Video enhancement aims to exploit sub-pixel information from contextual frames to improve the quality and resolution of videos, which primarily includes video frame interpolation, video super-resolution, and spatio-temporal video super-resolution.\\
\noindent \textbf{Video Frame Interpolation (VFI)} targets generating the intermediate frames between given consecutive inputs. Early CNN-based methods~\citep{VFI:Neighbor, VFI:depth, VFI:liu2017video, VFI:zhu2024dual, VFI:huang2022real} mainly rely on optical flow estimation or direct frame synthesis, but often suffer from limited receptive fields and poor handling of large motion. Therefore, Transformer-based approaches~\citep{VFI_trans_liu2024sparse, VFI_trans_park2023biformer, VFI:clearer} have been proposed to model long-range dependencies, significantly improving the quality and detail of generated video.\\
\noindent \textbf{Video Super-Resolution (VSR)} aims to reconstruct high-resolution (HR) video from low-resolution (LR) inputs. Early VSR methods primarily used explicit optical flow alignment~\citep{caballero2017real, tao2017detail, chan2021basicvsr}, dynamic filtering~\citep{jo2018deep}, deformable convolutions~\citep{wang2019edvr}, and temporal attention mechanisms~\citep{li2020mucan, xu2024enhancing}. With the increasing prominence of the Transformer's powerful representation capabilities, numerous Transformer-based VSR methods have been proposed, achieving progressive success. For example, PSRT~\citep{shi2022rethinking} leverages a multi-frame self-attention mechanism to jointly process features from the current input frame and the propagated features. MIA~\citep{zhou2024video} further boosts performance by leveraging masked intra-frame and inter-frame attention blocks to better use of previously enhanced features.\\
\noindent \textbf{Spatio-Temporal Video Super-Resolution (STVSR)} aims to simultaneously enhance spatial and temporal resolution, combining VSR and VFI, and presents greater challenges. Among the most representative real-time Transformer-based models is RSTT~\citep{STVSR:RSTT}, which achieves state-of-the-art performance by constructing feature dictionaries from different levels of encoders and repeatedly querying them during the decoding stage.\\
Although powerful transformer-based models have demonstrated superiority in enhancing spatial resolution and perceptual quality, their high computational cost hinders practical deployment. This paper is the first to propose an efficient model compression method specifically for video enhancement to facilitate its deployment.
\subsection{Model Quantization}
Model quantization aims to reduce the model's bit-width, from the Float 32-bit used in training to int 8, 4, or 2-bit for deployment, significantly reducing computational and memory costs and is widely applied in various fields such as LLM and VLM, etc. Quantization is divided into post-training quantization (PTQ) and Quantization-Aware Training (QAT). QAT, requiring simultaneous training and quantization, demands significant resources and data. PTQ, applied after pre-training, is more efficient and thus receives greater research focus.
Early PTQ methods focused on minimizing quantization error using efficient calibration techniques~\citep{yuan2022ptq4vit, li2023repq}. Recent approaches, such as AdaRound~\citep{wu2024adalog} and BRECQ~\citep{li2021brecq}, refine weight quantization by minimizing layer-wise output discrepancies. Additionally, robustness-oriented methods like NoisyQuant~\citep{liu2023noisyquant}, OASQ~\citep{ma2024outlier}, and ERQ~\citep{zhong2024erq} enhance PTQ performance by mitigating quantization noise, suppressing outliers, or optimizing error-aware objectives. However, existing work mainly addresses high-level vision/language tasks and is often unsuitable for pixel-level image/video enhancement, which is sensitive to quantization errors due to its reliance on fine-grained features. Recent research has thus started exploring quantization for pixel-level image enhancement and super-resolution~\citep{liuefficient,chen2025q,chen2024binarized,qin2023quantsr,zhang2024flexible}. For example, DBDC+Pac~\citep{DbdcPac} introduces a PTQ framework for image super-resolution, combining calibration techniques with knowledge distillation from a full-precision teacher model. Similarly, 2DQuant~\citep{2dquant} targets SwinIR~\citep{liang2021swinir} by proposing a one-sided search algorithm to quantize sensitive activations, such as post-softmax and post-GELU~\citep{hendrycks2016gaussian} layers. These methods often overlook inter-frame differences in multi-frame video enhancement, limiting detail representation and resulting in blurred images.
\begin{figure}[t]
\vspace{-4mm}
\centering
\includegraphics[width=1.0\textwidth]{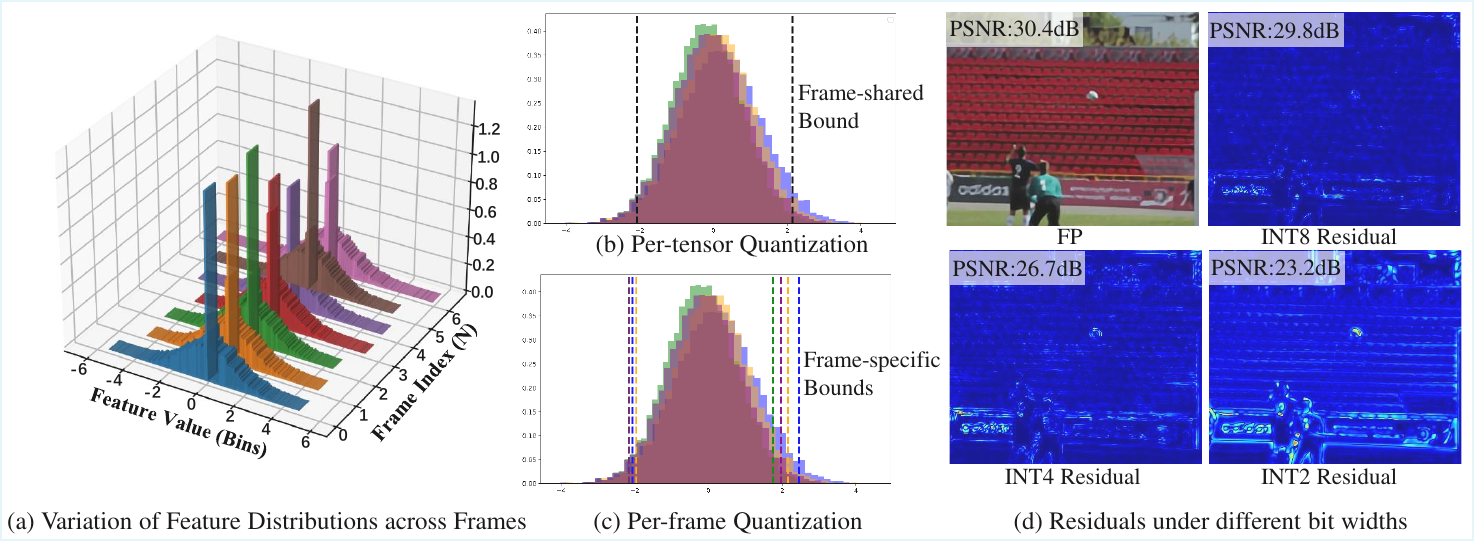}
\vspace{-4mm}
\caption{Finding and Motivation. (a) In multi-frame video enhancement, activation distributions vary significantly across frames. Traditional per-tensor quantization (b) fails to dynamically adjust quantization bounds for these variations, but our method (c) achieves this dynamic adjustment. (d) We calculated PSNR and residual maps for FP, INT8, INT4, and INT2 with respect to GT. The significant gap between low-bit (INT2/4) and full-precision (FP) model suggests that low-bit struggles to learn directly from FP. This inspired us to use multiple teacher models for supervision.}
\label{fig:fig1}
\vspace{-2mm}
\end{figure}

\section{Methodology}
\subsection{Problem Formulation}
Model quantization aims to learn appropriate clipping ranges~\citep{2dquant,DbdcPac,yuan2022ptq4vit} for each tensor (e.g., weights or activations) in order to minimize the discrepancy between the outputs of the full-precision model and the quantized model. Following previous work~\citep{2dquant,liu2025condiquant,DbdcPac}, we use fake quantization~\cite{jacob2018quantization} to simulate the quantization
process. Given a pre-learned clipping range $[lb, ub]$ for a tensor $x$, the quantization and dequantization process is defined as follows:
\begin{equation}
x_\text{clip} = \text{clamp}(x, lb, ub) = \min(\max(x, lb), ub),
\end{equation}
\begin{equation}
x_\text{int} = \text{round} \left( \frac{x_\text{clip} - lb}{\Delta} \right), \quad \Delta = \frac{ub - lb}{2^N - 1}, \quad\hat{x} = x_\text{int} \cdot \Delta + lb,
\end{equation}
Where $x_\text{clip}$ is the clipped tensor, $x_\text{int} \in \{0, 1, \dots, 2^N - 1\}$ is the quantized integer, $\Delta$ is the quantization step size, and $\hat{x}$ is the dequantized approximation of the original value. Linear and MatMul layers are the most computationally intensive components in Transformer-based architectures. We follow existing work~\citep{2dquant,liu2023noisyquant} to focus quantization on these modules. Since the quantization function is non-differentiable due to rounding, we also adopt the Straight-Through Estimator (STE)~\cite{courbariaux2016binarized} during training to approximate gradients and enable end-to-end optimization under quantized settings. More details are in the Appendix. 
\subsection{Observations and Motivation}
\noindent\textbf{Observation 1: Inability to allocate varying representational capacity across frames.} 
\label{sec:observation1}Previous studies have thoroughly examined the statistical properties of activations in Transformer-based architectures, uncovering long-tailed distributions and a mix of symmetric and asymmetric behaviors across 
layers~\cite{2dquant,yuan2022ptq4vit,li2023repq,moon2024instance,zhong2024erq}. However, these analyses are largely confined to single-frame scenarios. In the context of quantizing multi-frame video enhancement networks, it is essential to preserve the capability of the full-precision network to effectively integrate inter-frame texture information and motion cues—a challenge that vanilla methods fail to address.\\
Our analysis reveals that multi-frame networks perceive and process each frame in the input tensor differently. As shown in Figure~\ref{fig:fig1}(a), we collect per-frame activation statistics and observe significant disparities in activation distributions across frames. In particular, the value ranges (i.e., the minimum and maximum activation values) vary considerably, indicating that the network allocates representational capacity unevenly across frames.
These differences are influenced by the model’s frame-dependent attention dynamics. In particular, the network assigns different attention weights to each frame, resulting in varied activation distributions.
Consequently, applying existing Transformer quantization methods~\cite{2dquant,yuan2022ptq4vit,li2023repq,moon2024instance,zhong2024erq}, which typically assume a unified activation distribution, can be suboptimal for multi-frame models. Such methods overlook inter-frame variation, leading to inefficient quantization and increased error, ultimately degrading the quality of the output.\\
\noindent\textbf{Observation 2: Over-reliance on full-precision teachers limits low-bit model learning.} Low-bit quantization inevitably reduces the representational capacity of models, which brings significant challenges to multi-frame video enhancement tasks. Under low-bit quantization, both activation values and network precision degrade noticeably, making it difficult to maintain the clear motion trajectories and rich texture details required for these tasks. As shown in Figure~\ref{fig:fig1}(c), directly quantizing the network to 4-bit/2-bit and applying it without adaptation leads to severe artifacts and motion blur. The main reason for this degradation lies in the large quantization errors introduced when networks are directly quantized to low-bit precision. Furthermore, traditional methods often overlook the capacity gap between the teacher model (FP32) and the student model (e.g., 4-bit or 2-bit), relying solely on full-precision teachers for knowledge distillation~\citep{2dquant,DbdcPac,gou2021knowledge}. This makes it difficult for the student model to learn high-quality mappings within its limited capacity, further increasing the challenge of achieving satisfactory performance under low-bit constraints.
\begin{figure}[t]
\centering
\vspace{-3mm}
\includegraphics[width=1.0\textwidth]{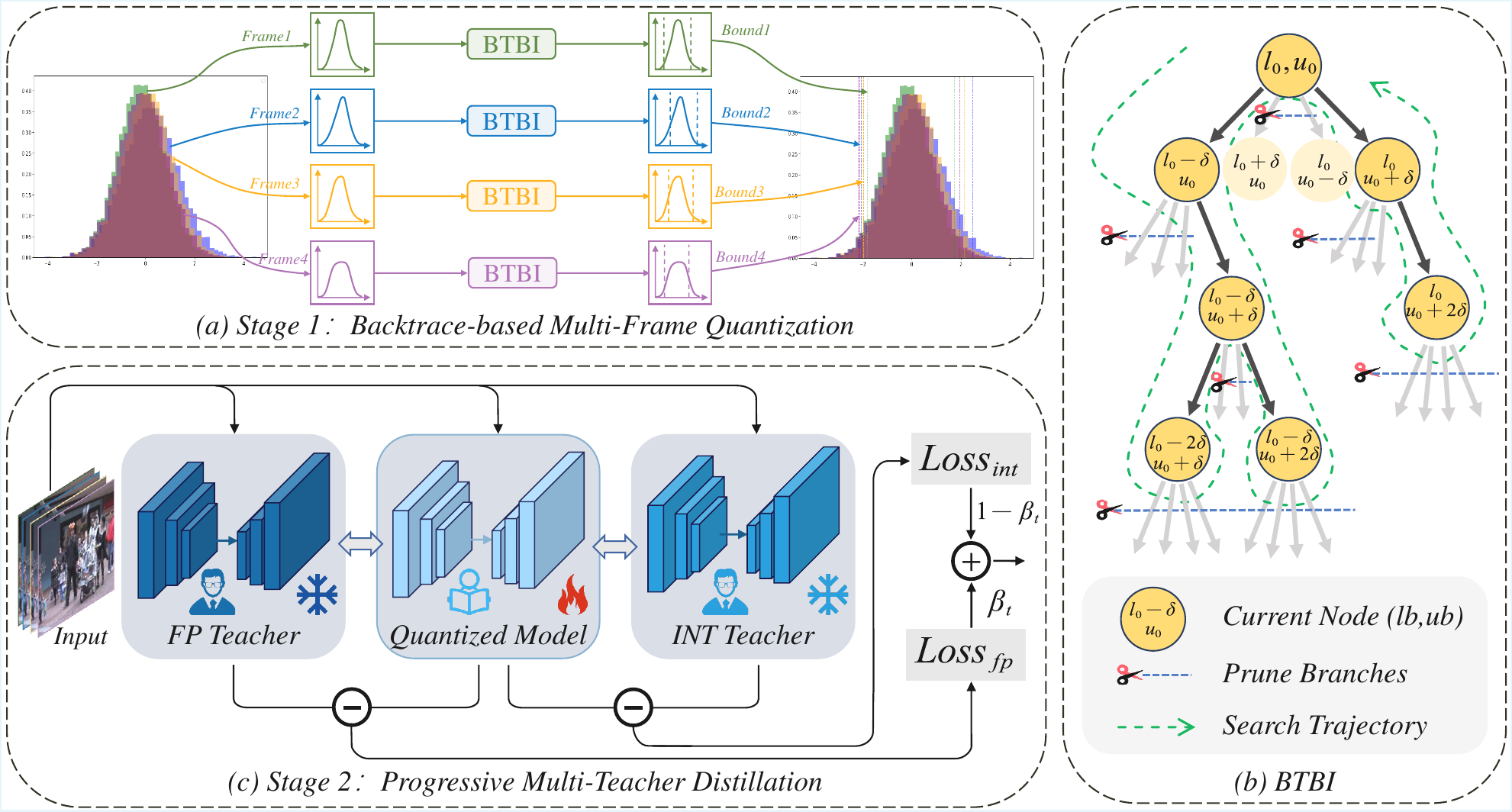}
\vspace{-4mm}
\caption{The overall framework of our proposed method.}
\vspace{-4mm}
\label{fig:framework}
\end{figure}
\subsection{Proposed Method}
Based on the above observations, we propose a two-stage quantization framework. The coarse stage uses Backtracking-based Multi-Frame Quantization (BMFQ) to initialize asymmetric bounds efficiently. The fine stage applies Progressive Multi-Teacher Distillation (PMTD) to refine bounds.
\subsubsection{Backtracking-based Multi-Frame Quantization}
Motivated by Observation 1, we adopt a per-frame quantization strategy to handle frame-wise variations in activation distributions and representational capacity. Given a multi-frame activation tensor $X \in \mathbb{R}^{N \times C \times H \times W}$, where $N$ denotes the number of frames, we independently search the clipping bounds for each frame  $X_i = X[i, :, :, :]$, yielding frame-specific clipping parameters $(lb_i, ub_i)$ for $i = 1, \dots, N$. This strategy enables the quantizer to adapt to frame-specific activation statistics, thereby improving quantization accuracy.\\
To robustly estimate the clipping bounds during the coarse stage, we formulate the selection of $(lb_i, ub_i)$ as a constrained optimization problem that minimizes quantization-induced distortion over a percentile-based search space $S_i$ derived from the empirical distribution of $X_i$:
\begin{equation}
    (lb_i^*, ub_i^*) = \argmin_{(lb, ub) \in S_i} \; \mathbb{E}_{x \sim X_i} \left[(x - Q_{lb, ub}(x))^2\right],
\label{eq:clip_opt}
\end{equation}
\begin{minipage}[t]{0.52\textwidth}
    Here, $Q_{lb, ub}(\cdot)$ denotes a uniform quantizer~\citep{liu2022nonuniform} with clipping range $[lb, ub]$. To mitigate the influence of outliers, we constrain the search space $S_i$ using percentiles: $lb \in [p_{0.1}(X_i), p_{10}(X_i)]$, $ub \in [p_{90}(X_i), p_{99.9}(X_i)]$, where $p_k(X)$ denotes the $k$-th percentile of the tensor $X$.
    To efficiently solve the optimization problem in Eq.~\eqref{eq:clip_opt}, we introduce a Backtracking-based Bound Initialization (BTBI) algorithm. Starting from an initial estimate derived from the percentiles of $X_i$, the algorithm recursively explores candidate bounds by adjusting $lb_i$ and $ub_i$ within the search space ${S_i}$. At each step, it evaluates the quantization error and updates the optimal bounds if a better configuration is found. To avoid redundant searches, previously visited bounds are skipped. The algorithm backtracks to explore alternative adjustments when no further improvement is achieved, terminating when all candidates are evaluated or a convergence threshold is met.
    In contrast to traditional methods that uniformly shrink bounds~\citep{2dquant} or adjust them sequentially~\citep{DbdcPac}, BTBI is less sensitive to outliers and explores a richer set of candidate configurations. By combining frame-wise adaptation with   recursive backtracking search, BTBI robustly converges to optimal clipping parameters for each frame. To enhance understanding, we provide a detailed algorithm of our BTBI in Algorithm~\ref{alg:backtrack}.
\end{minipage}
\hfill
\begin{minipage}[t]{0.45\textwidth}
    \vspace{-4mm}
    \begin{algorithm}[H]
    \caption{Backtracking-based Bound Initialization (BTBI) pipeline}
    \label{alg:backtrack}
    \KwIn{$X$, step sizes $\Delta L$, $\Delta U$, threshold $\varepsilon$}
    \KwOut{Optimal bounds $lb^*, ub^*$}
    \DontPrintSemicolon
    \BlankLine
    $visited \leftarrow \emptyset$, $error_{\min} \leftarrow \infty$\;
    \BlankLine
    \SetKwFunction{FBacktrack}{Backtrack}
    \SetKwProg{Fn}{Function}{:}{\KwRet}
    \Fn{\FBacktrack{$lb, ub$}}{
        \If{$(lb, ub) \in visited$ \textbf{or} out-of-range}{
            \KwRet\;
        }
        $visited \leftarrow visited \cup \{(lb, ub)\}$\;
        $X_q \leftarrow$ Quantize $X$ using $(lb, ub)$\;
        $err \leftarrow \|X - X_q\|_2$\;
        \If{$err > error_{\min} + \varepsilon$}{
            \KwRet\;
        }
        \If{$err < error_{\min}$}{
            $error_{\min} \leftarrow err$, $lb^* \leftarrow lb$, $ub^* \leftarrow ub$\;
        }
        \ForEach{$(\delta_l, \delta_u) \in \{\pm \Delta L, \pm \Delta U\}$}{
            \FBacktrack{$lb + \delta_l, ub + \delta_u$}\;
        }
    }
    \BlankLine
    \textbf{\FBacktrack}($lb_0$, $ub_0$)\;
    \Return $lb^*, ub^*$\;
    \end{algorithm}
\end{minipage}

\subsubsection{Progressive Multi-Teacher Distillation}
As revealed by Observation 2, training accurate quantized models under extremely low-bit settings (e.g., 4-bit or 2-bit) remains challenging due to limited capacity and optimization instability. To address this, we propose Progressive Multi-Teacher Distillation (PMTD), a hierarchical distillation framework that leverages both high-bit and full-precision teachers to facilitate low-bit training. Instead of directly distilling knowledge from a full-precision (FP) teacher to a low-bit student, which often suffers from large representational gaps, PMTD introduces intermediate-bit teacher models (e.g., 8-bit) between the full-precision (FP) teacher and the low-bit student. These intermediate teachers serve as quantization-aware approximations of the FP model, providing smoother supervision and easing the optimization process. This hierarchical approach effectively bridges the representational gap between student and teacher models, ensuring more stable training dynamics and improved performance under extreme quantization constraints.
Specifically, to train a 4-bit quantized model, PMTD first uses the full-precision model as a teacher to train an 8-bit model. 
When training the 4-bit model, both the full-precision network and the 8-bit network are used as teacher models, as illustrated in Figure \ref{fig:framework}(c).
The distillation process is formally defined by the following loss function:
\begin{equation}
\mathcal{L}_{\text{PMTD}} =  (\mathcal{L}_{\text{INT}} + \alpha(t) \cdot \mathcal{L}_{\text{FP}})/(1+\alpha(t)),
\label{eq:pmt_loss}
\end{equation}
where $\mathcal{L}_{\text{INT}}$ denotes the total loss from intermediate-bit teachers (e.g., 8-bit), and $\mathcal{L}_{\text{FP}}$ represents the loss from the full-precision teacher. The balancing coefficient $\alpha(t)$ linearly increases over time and is defined as $\alpha(t) = \min\left(1, \frac{t}{T_{\text{warmup}}}\right)$, where $T_{\text{warmup}}$ is a hyperparameter controlling the warm-up duration.
Each teacher-specific loss consists of two components: an output-level reconstruction loss and an intermediate feature-matching loss:
\begin{align}
\mathcal{L}_{\text{INT}} &= \sum_{k=1}^{K} \left( \mathcal{L}_{\text{rec}}^{(k)} + \lambda \cdot \mathcal{L}_{\text{feat}}^{(k)} \right), \label{eq:int_loss} \\
\mathcal{L}_{\text{FP}} &= \mathcal{L}_{\text{rec}}^{\text{FP}} + \lambda \cdot \mathcal{L}_{\text{feat}}^{\text{FP}}, \label{eq:fp_loss}
\end{align}
where $K$ is the number of intermediate-bit teachers (e.g., $K=2$ when using 4-bit and 8-bit teachers), $\mathcal{L}_{\text{rec}}$ is the $\ell_2$ loss~\citep{lim2017enhanced} between the student and teacher outputs, and $\mathcal{L}_{\text{feat}}$ is the mean squared error (MSE)~\citep{bauer1999empirical} between selected intermediate feature representations. The balancing coefficient $\lambda$ is set to 5 to emphasize the importance of internal consistency.
By gradually transitioning supervision from intermediate-bit to full-precision teachers, PMTD effectively reduces the training difficulty of low-bit models, mitigates quantization errors, and offers a more stable optimization path. This hierarchical approach ensures high-quality quantized outputs, even under extreme quantization constraints.
\begin{table*}[t]
\caption{Quantitative comparison of different methods on four STVSR benchmarks. The best and the second best results are in \best{bold} and \second{bold}.}
\label{tab:main_table}
\centering
\renewcommand{\arraystretch}{1.2}
\setlength{\tabcolsep}{4pt}
\scalebox{0.90}{
\begin{tabular}{l|c|cc|cc|cc|cc}
\toprule
\multirow{2}{*}{Method} & \multirow{2}{*}{Bit} &
\multicolumn{2}{c|}{Vid4} &
\multicolumn{2}{c|}{Vimeo-Fast} &
\multicolumn{2}{c|}{Vimeo-Medium} &
\multicolumn{2}{c}{Vimeo-Slow} \\
 &  &
PSNR↑ & SSIM↑ &
PSNR↑ & SSIM↑ &
PSNR↑ & SSIM↑ &
PSNR↑ & SSIM↑ \\
\midrule
RSTT-S~\cite{STVSR:RSTT} & 32/32 & 26.29 & 0.7941 & 36.58 & 0.9381 & 35.43 & 0.9358 & 33.30 & 0.9123 \\
Trilinear & 32/32 & 22.90 & 0.5883 & 29.45 & 0.8019 & 29.16 & 0.8351 & 28.21 & 0.8091 \\
\midrule
OpenVINO~\citep{openvino} & 2/2 &18.24  &0.3151  &23.39  &0.6103  &23.60  &0.6227  &23.87  &0.6242  \\
TensorRT~\citep{tensorrt} & 2/2 &20.31  &0.5118  &23.41  &0.6106  &23.61  &0.6236  &23.88  &0.6283  \\
SNPE~\citep{snpe} & 2/2 &15.22  &0.2378 &23.40  &0.6106  &23.61  &0.6241  &23.88  &0.6281  \\
Percentile~\citep{Percentile} & 2/2 &12.67  &0.1349  &15.27  & 0.2274 &14.80  &0.2165  &14.67  &0.2156    \\
MinMax~\citep{MinMax} & 2/2 &10.34  &0.0138  &10.52  &0.0266  &10.48  &0.0289  &10.45  &0.0303  \\
NoisyQuant~\citep{liu2023noisyquant} & 2/2 &12.06  &0.1028  & 12.50 & 0.1669 & 11.81 & 0.1465 &11.49  & 0.1508 \\
DBDC+Pac~\citep{DbdcPac} & 2/2 &22.64  &0.5695  &28.94  &0.8254  &28.87  &0.8214  &27.86  &0.7905  \\
2DQuant~\citep{2dquant}  & 2/2 &\second{22.91}  &\second{0.5883}  &\second{29.38}  &\second{0.8315}  &\second{29.14}  &\second{0.8330}  &\second{28.18}  &\second{0.8086}  \\
\textbf{Ours} & 2/2 &\best{23.48}  &\best{0.6252}   &\best{30.33}   &\best{0.8424}   &\best{30.19}   &\best{0.8523}   &\best{29.14}   &\best{0.8316}   \\
\midrule
OpenVINO~\citep{openvino} & 4/4 & 18.84 & 0.4591 & 21.82 & 0.6372 & 21.63 & 0.6315 & 18.84 & 0.4591 \\
TensorRT~\citep{tensorrt} & 4/4 & 18.63 & 0.4578 & 21.74 & 0.6019 & 21.70 & 0.6324 & 18.63 & 0.4578 \\
SNPE~\citep{snpe} & 4/4 & 17.84 & 0.3977 & 21.64 & 0.6018 & 21.56 & 0.6301 & 18.76 & 0.4584 \\
Percentile~\citep{Percentile} & 4/4 & 23.26 & 0.6314 & 27.12 & 0.7664 & 27.16 & 0.7709 & 26.58 & 0.7531 \\
MinMax~\citep{MinMax} & 4/4 & 21.60 & 0.5242 & 26.41 & 0.6990 & 25.94 & 0.7059 & 25.44 & 0.6957 \\
NoisyQuant~\citep{liu2023noisyquant} & 4/4 &24.26 & 0.6905  & 31.22 & 0.8719 & 30.64 & 0.8705 & 29.61 & 0.8462 \\
DBDC+Pac~\citep{DbdcPac} & 4/4 & 24.50 & 0.6923 & 32.64 & 0.8857 & 32.06 & 0.8866 & 30.64 & 0.8643 \\
2DQuant~\citep{2dquant} & 4/4 & \second{25.04} & \second{0.7256} & \second{33.59} & \second{0.9035} & \second{32.83} & \second{0.9009} & \second{31.21} & \second{0.8766} \\
\textbf{Ours} & 4/4 & \best{25.42} & \best{0.7501} & \best{34.69} & \best{0.9181} & \best{33.74} & \best{0.9150} & \best{31.94} & \best{0.8903} \\

\bottomrule
\end{tabular}
}
\end{table*}
\section{Experiment and Analysis}
\subsection{Experiment Setting}
\noindent \textbf{Datasets and backbone.} We evaluate our method on three representative video enhancement tasks: Space-Time Video Super-Resolution (STVSR), Video Super-Resolution (VSR), and Video Frame Interpolation (VFI). For each task, we select state-of-the-art and popular methods as backbones: RSTT~\citep{STVSR:RSTT} for STVSR, MIA~\citep{zhou2024video} for VSR, and EMA-VFI~\citep{VFI:clearer} for VFI. The Vimeo-90K~\citep{STVSR:Vimeo90k} dataset is used for training across all tasks, with Vid4~\citep{STVSR:Vid4} and the Vimeo-90K test set serving as evaluation benchmarks. More details of data preparation and setting are provided in the supplementary material.\\
\noindent \textbf{Evaluation metrics.}  We use PSNR and SSIM~\citep{common:ssim} as evaluation metrics, computed on the luminance (Y) channel of the YCbCr color space. To further evaluate perception-oriented metrics, LPIPS~\citep{zhang2018unreasonable} and NIQE~\citep{mittal2012making} are used to assess the perceptual quality of videos.\\
\noindent \textbf{Implementation details.} We adopt the Adam optimizer~\citep{common:adam} with an initial learning rate of $2 \times 10^{-4}$ and apply Cosine Annealing~\citep{common:cosineAnnealing} over 20,000 iterations. The batch size is set to 8 and 2 per GPU during the initialization and distillation-based fine-tuning phases, respectively. Random cropping, rotations, and flipping are applied to enhance training robustness. All experiments are implemented in Python with PyTorch~\citep{common:pytorch} and conducted on 8 NVIDIA V100 GPUs. 
\subsection{Quantitative Results}
Table~\ref{tab:main_table} presents quantitative comparisons of various methods under 2/2, 4/4, bit-width across four STVSR benchmarks. Traditional quantization approaches, such as OpenVINO~\citep{openvino} and TensorRT~\citep{tensorrt}, face challenges in pixel-level video enhancement, resulting in model performance scores of only 18.24dB and 20.31dB on Vid4 at 2-bit quantization. Although DBDC+Pac~\citep{DbdcPac} and 2DQuant~\citep{2dquant} are tailored for low-level vision tasks with enhanced sharpness awareness, which somewhat mitigates the performance drop due to quantization, they still lag behind our proposed method. This is primarily due to their limitations in managing multi-frame distribution differences and detail enhancement. Our method achieves the best performance across all scenarios and benchmarks, notably surpassing existing methods by nearly 1 dB on the Vimeo benchmark, underscoring the effectiveness of our approach. In a similar manner, our method demonstrates superior performance across all benchmarks and bit-width configurations for both Video Super-Resolution (VSR) and Video Frame Interpolation (VFI) tasks. As detailed in Tables~\ref{tab:main_table_vfi} and~\ref{tab:main_table_vsr_compact_final}, our approach exemplifies remarkable generalization capabilities, consistently outperforming existing methods in all benchmarks and bits. More results on additional bit-width settings and benchmarks can be found in Appendix.
\begin{table*}[t]
\centering
\caption{Quantitative comparison of different methods on two VFI EMA-VFI variants ([T] and [D]), evaluated on the Vimeo90K benchmark under 4-bit.}
\label{tab:main_table_vfi}
\renewcommand{\arraystretch}{1.15}
\setlength{\tabcolsep}{4pt}
\scalebox{0.90}{
\begin{tabular}{l c | cccc | cccc}
\toprule
\multirow{2}{*}{Method} & \multirow{2}{*}{Bit} & \multicolumn{4}{c|}{EMA-VFI [T]~\citep{VFI:ema-vfi}} & \multicolumn{4}{c}{EMA-VFI [D]~\citep{VFI:clearer}} \\
\cmidrule(lr){3-6} \cmidrule(lr){7-10}
& & PSNR$\uparrow$ & SSIM$\uparrow$ & LPIPS$\downarrow$ & NIQE$\downarrow$ & PSNR$\uparrow$ & SSIM$\uparrow$ & LPIPS$\downarrow$ & NIQE$\downarrow$ \\
\midrule
Baseline  & 32/32 & 29.41 & 0.9279 & 0.086 & 6.736 & {30.29} & {0.9418} & 0.078 & 6.545 \\
\midrule
OpenVINO~\citep{openvino}        & 4/4   & 26.03 & 0.8703 & 0.222 & 8.022 & 25.38 & 0.8579 & 0.257 & 8.2784 \\
TensorRT~\citep{tensorrt}        & 4/4   & 25.33 & 0.8551 & 0.268 & 8.582 & 25.21 & 0.8537 & 0.269 & 8.4035 \\
SNPE~\citep{snpe}        & 4/4   & 25.49 & 0.8581 & 0.259 & 8.500 & 25.83 & 0.8683 & 0.236 & 8.0399 \\ 
Percentile~\citep{Percentile}      & 4/4   & 26.82 & 0.8919 & 0.185 & 7.765 & 28.54 & 0.9198 & 0.132 & 7.0667 \\ 
MinMax~\citep{MinMax}          & 4/4   & 23.03 & 0.7918 & 0.389 & 9.475 & 24.19 & 0.8309 & 0.313 & 8.5153 \\ 
DBDC+Pac\citep{DbdcPac}        & 4/4   & 27.30 & 0.8976 & 0.171 & 7.545 & 28.36 & 0.9179 & 0.134 & 7.1221 \\
2DQuant~\citep{2dquant}         & 4/4   & \second{28.06} & \second{0.9110} & \second{0.152} & \second{7.494} & \second{28.78} & \second{0.9233} & \second{0.120} & \second{6.9884} \\
Ours            & 4/4   & \best{28.41} & \best{0.9162} & \best{0.136} & \best{7.361} & \best{29.59} & \best{0.9335} & \best{0.101} & \best{6.7881} \\
\bottomrule
\end{tabular}
}
\end{table*}
\begin{table*}[t]
\centering
\caption{Quantitative comparison of different methods on two VSR benchmarks under 4-bit.}
\label{tab:main_table_vsr_compact_final}
\renewcommand{\arraystretch}{1.0} 
\setlength{\tabcolsep}{3pt} 
\scalebox{0.90}{ 
\begin{tabular}{l|l|c|c|c|c|c|c|c|c}
  \toprule
  Benchmark & Metric 
  & \makecell{Baseline:\\MIA~\citep{zhou2024video}} 
  & \makecell{TensorRT\\~\citep{tensorrt}} 
  & \makecell{SNPE\\~\citep{snpe}} 
  & \makecell{Percentile\\~\citep{Percentile}} 
  & \makecell{MinMax\\~\citep{MinMax}} 
  & \makecell{DBDC\\+Pac~\citep{DbdcPac}} 
  & \makecell{2DQuant\\~\citep{2dquant}} 
  & Ours \\
  \midrule
  \multirow{2}{*}{Vimeo90K}
  & PSNR↑ & 38.32 & 31.81 & 32.42 & 35.15 & 34.13 & 36.64 & \second{36.92} & \best{37.34} \\
  & SSIM↑ & 0.9532 & 0.8612 & 0.8805 & 0.9262 & 0.9119 & 0.9404 & \second{0.9434} & \best{0.9487} \\
  \midrule
  \multirow{2}{*}{Vid4} 
  & PSNR↑ & 28.20 & 24.48 & 24.22 & 26.14 & 25.60 & 27.26 & \second{27.38} & \best{27.64} \\
  & SSIM↑ & 0.8507 & 0.6758 & 0.6877 & 0.7805 & 0.7494 & 0.8230 & \second{0.8287} & \best{0.8341} \\
  \bottomrule
\end{tabular}
}
\end{table*}
\subsection{Qualitative Results}
\begin{figure}[t!]
    \centering
    \includegraphics[width=1.0\textwidth]{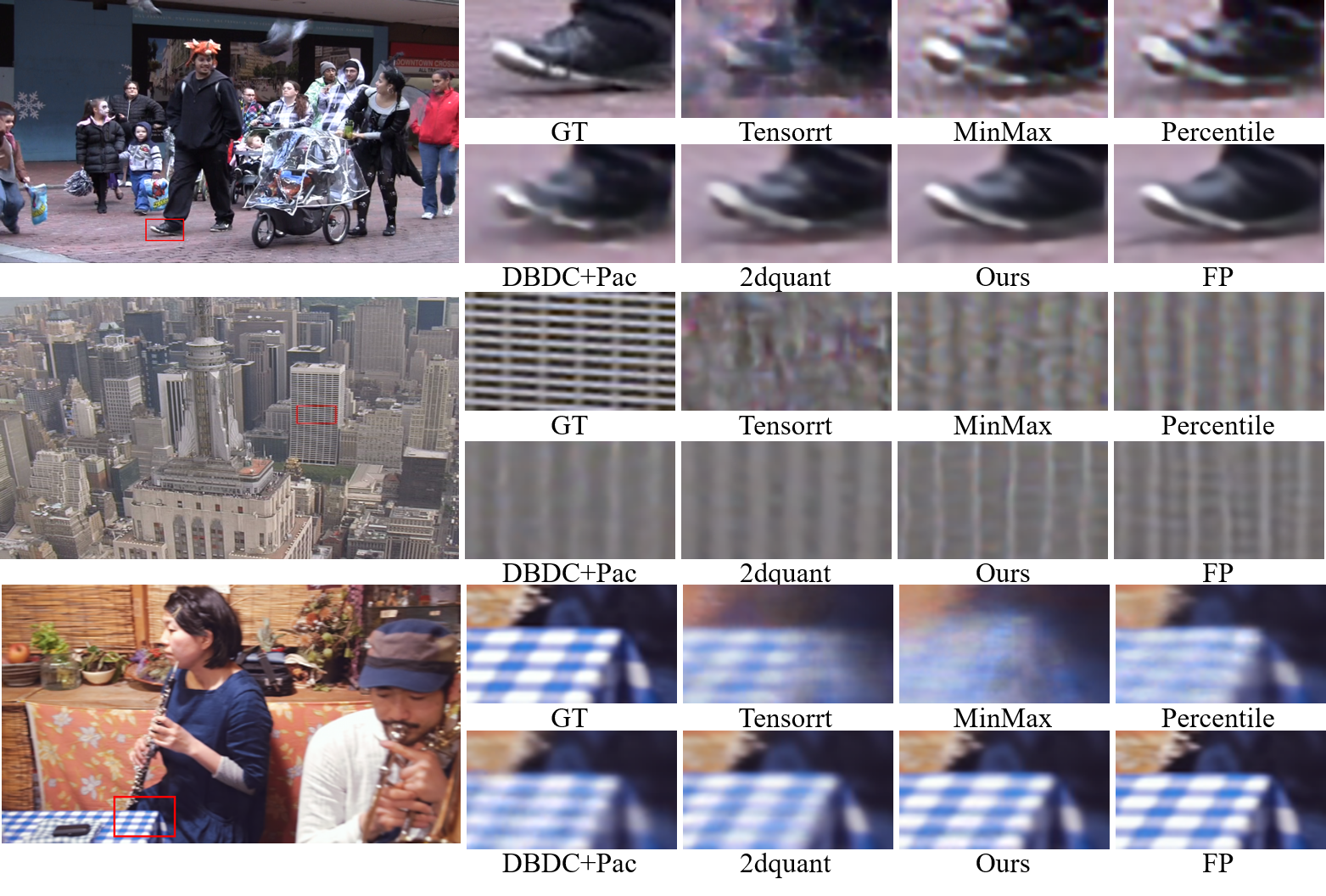}
    \vspace{-4mm}
\caption{Visual comparisons under 4-bit quantization for three video enhancement tasks: from top to bottom are STVSR, VSR, and VFI tasks. More results are provided in the Appendix.}
    \vspace{-4mm}   
    \label{fig:results}
    
\end{figure}

To verify the visual quality, we present the visual comparisons of different PTQ methods applied to STVSR, VSR and VFI tasks under 4-bit quantization, as shown in Figure~\ref{fig:results}. Traditional methods such as MinMax~\citep{MinMax} and Percentile~\citep{Percentile} exhibit noticeable artifacts, while other methods like DBDC+Pac~\citep{DbdcPac} and 2dquant~\citep{2dquant} suffer from detail blurring issues, particularly in complex scenes. However, our proposed method consistently produces sharper edges and more faithful textures that are visually closer to the full-precision outputs. In more challenging cases, such as the Vimeo-Fast dataset where motion and fine details coexist, our proposed method better preserves structural information and avoids common artifacts. This highlights the visual superiority of our approach. More visual effects and comparisons with user studies will be presented in the Appendix.

\subsection{Ablation Studies}
To validate the effectiveness of the proposed core idea, we design several ablation experiments to explore the multi-teacher distillation strategy the frame-wise quantization strategy.
\begin{minipage}[t]{0.58\textwidth}Specifically, we conduct experiments on STVSR in a 2-bit compression setting, removing these core modules one by one, with results shown in Table~\ref{tab:ablation}. It is seen that the baseline without any core ideas achieves only 12.67dB. By introducing the frame-wise quantization strategy, the network perceives differences between frames, improving performance. Furthermore, the BMFQ helps the network adaptively learn clipping ranges for each frame, boosting 
\end{minipage}
\hfill
\begin{minipage}[t]{0.40\textwidth}
\centering
\vspace{-1mm}
    \captionof{table}{Ablation studies.}
    \vspace{-2mm}

    \label{tab:ablation}
    \setlength{\tabcolsep}{1pt}
    \footnotesize
    \begin{tabular}{ccc>{\centering\arraybackslash}p{2cm}}
        \toprule
        \makecell{Per-Frame \\Quantization} &\makecell{BMFQ}  &\makecell{PMTD}  & \makecell{PSNR$\uparrow$} \\
        \midrule
        \ding{55} & \ding{55} & \ding{55} & 12.67   \\
        \checkmark & \ding{55} &  \ding{55} & 19.64      \\
        \checkmark  & \checkmark & \ding{55}   & 27.56        \\
        \checkmark  & \checkmark  & \checkmark  & 30.33    \\
        \bottomrule
    \end{tabular}

\end{minipage}
 model performance to 27.56dB. Finally, with the introduction of the multi-teacher distillation strategy, the low-bit network learns prior knowledge from different teachers, further improving model performance despite limited capacity. This validates the effectiveness of the proposed core modules. 
More ablation studies are presented in the Appendix.
\section{Conclusion}
\label{sec:conclusion}
We introduced a novel coarse-to-fine PMQ-VE, addressing key challenges in quantizing multi-frame video enhancement models. BMFQ is proposed to establish robust quantization bounds through a percentile-based initialization and backtracking search, ensuring efficient quantization across frames. PMTD enhances the quality of low-bit models by utilizing a progressive distillation strategy with both full-precision and quantized teachers, bridging the gap between high-precision and low-bit models. Experiments on STVSR, VSR, and VFI tasks show that our PMQ-VE achieves state-of-the-art performance and visually pleasing results. 
Limitation and Future Work: Although our PMQ-VE has achieved promising results on Transformer-based video enhancement methods, more diffusion-based Transformer (DiT) methods can be tested. In the future, we plan to extend our method to more video enhancement tasks and models to facilitate the deployment of video models in the community.



\clearpage
\bibliographystyle{plainnat} 
\bibliography{references}    
\end{document}